\documentclass[journal]{IEEEtran}

\usepackage[dvipsnames]{xcolor}
\usepackage{xcolor,soul,framed} 

\colorlet{shadecolor}{yellow}
\usepackage[pdftex]{graphicx}
\graphicspath{{../pdf/}{../jpeg/}}
\DeclareGraphicsExtensions{.pdf,.jpeg,.png}

\usepackage[cmex10]{amsmath}
\usepackage{array}
\usepackage{mdwmath}
\usepackage{mdwtab}
\usepackage{eqparbox}
\usepackage{url}
\usepackage{amsthm}
\newtheorem{definition}{Definition}
\usepackage{makecell}

\begin{document}
\bstctlcite{IEEEexample:BSTcontrol}
    \title{Public Health Informatics: Proposing Causal Sequence of Death Using Neural Machine Translation}
  \author{Yuanda Zhu *,~\IEEEmembership{Student Member,~IEEE,}
      Ying Sha *,~\IEEEmembership{Student Member,~IEEE,}
      Hang Wu *,~\IEEEmembership{Student Member,~IEEE,}\\
      Mai Li,~\IEEEmembership{Student Member,~IEEE,}
      Ryan A. Hoffman,~\IEEEmembership{Student Member,~IEEE,}
      May D. Wang,~\IEEEmembership{Senior Member,~IEEE}

  \thanks{The work was supported by the U.S. Department of Health and Human Services Centers for Disease Control and Prevention under Award HHSD2002015F62550B, the NIH National Center for Advancing Translational Sciences under Award UL1TR000454, and the National Science Foundation Award NSF1651360.}
  \thanks{Y. Zhu is with School of Electrical and Computer Engineering, Georgia Institute of Technology, Atlanta, GA 30332 USA (e-mail: yzhu94@gatech.edu).}
  \thanks{Y. Sha is with School of Biology, Georgia Institute of Technology, Atlanta, GA 30332 USA (e-mail: ysha8@gatech.edu).}
  \thanks{H. Wu is with Department of Biomedical Engineering, Georgia Institute of Technology, Atlanta, GA 30332 USA (e-mail: hangwu@gatech.edu)}%
  \thanks{M. Li is with Department of Electronic Science and Technology, University of Science and Technology of China, Hefei, Anhui Province, China (e-mail: lm333@mail.ustc.edu.cn).}%
  \thanks{R. A. Hoffman and M. D. Wang are with Department of Biomedical Engineering, Georgia Institute of Technology and Emory University, Atlanta, GA 30332 USA (phone: 404-385-2954; e-mail: rhoffman12@gatech.edu, maywang@gatech.edu)}
  \thanks{*The first three authors contributed equally to this work.}}


\maketitle

\begin{abstract}

Each year there are nearly 57 million deaths around the world, with over 2.7 million in the United States. Timely, accurate and complete death reporting is critical in public health, as institutions and government agencies rely on death reports to analyze vital statistics and to formulate responses to communicable diseases. Inaccurate death reporting may result in potential misdirection of public health policies. Determining the causes of death is, nevertheless, challenging even for experienced physicians. To facilitate physicians in accurately reporting causes of death, we present an advanced AI approach to determine a chronically ordered sequence of clinical conditions that lead to death, based on decedent’s last hospital discharge record. The sequence of clinical codes on the death report is named as causal chain of death, coded in the tenth revision of International Statistical Classification of Diseases (ICD-10); in line with the ICD-9-CM Official Guidelines for Coding and Reporting, the priority-ordered clinical conditions on the discharge record are coded in ICD-9. We identify three challenges in proposing the causal chain of death: two versions of coding system in clinical codes, medical domain knowledge conflict, and data interoperability. To overcome the first challenge in this sequence-to-sequence problem, we apply neural machine translation models to generate target sequence. Along with three accuracy metrics, we evaluate the quality of generated sequences with the BLEU (BiLingual Evaluation Understudy) score and achieve 16.04 out of 100. To address the second challenge, we incorporate expert-verified medical domain knowledge as constraint in generating output sequence to exclude infeasible causal chains. Lastly, we demonstrate the usability of our work in a Fast Healthcare Interoperability Resources (FHIR) interface to address the third challenge.

\end{abstract}


%
\IEEEpeerreviewmaketitle


\section{Introduction}

\IEEEPARstart{T}{here} are more than 2.7 million deaths happening in the United States every year \cite{xu2018deaths}, with nearly 57 million deaths per year around the world \footnote{https://www.who.int/news-room/fact-sheets/detail/the-top-10-causes-of-death}. Accurate death reporting is essential for public health institutions such as the National Center for Health Statistics (NCHS) and the Centers for Disease Control and Prevention (CDC) to analyze vital statistics such as life expectancy and to formulate responses to communicable disease threats and epidemics. 
In addition to reporting simple information such as demographics of the deceased, an important component of death reporting is to determine the causes of deaths.
The U.S. death reporting system requires two types of causes of deaths to be filled on death certificates: a \emph{single} medical condition that is the underlying cause of death, and also an ordered list, a \emph{causal chain}, of medical conditions that lead to the death.

A causal chain of death consists one underlying cause of death, and zero, one or more immediate causes of death. The immediate causes of death are typically caused by the underlying cause of death. An example causal chain of death is ``chronic obstructive pulmonary disease, unspecified (ICD10: J44.9) $\rightarrow$ other disorders of lung (ICD10: J98.4)''. Here ICD10 stands for ``10th revision of the International Statistical Classification of Diseases and Related Health Problems'', a common coding system used in death reporting \footnote{https://www.cdc.gov/nchs/icd/icd10cm.htm}.

The process of determining such causal chains of death, nevertheless, is challenging, even for an experienced physician. Such a process involves careful reasoning with one's medical domain knowledge, provoking challenges for young or inexperienced physicians. Even worse, a sudden and unexpected death might further exacerbate the process of filling the death report when the physician could only find limited electronic health records of the deceased.

Complete and accurate reporting of the full chain of causes of death has multiple benefits. This data is an invaluable public health resource for tracking the prevalence of causes of death, targeting public health interventions, and tracking the effectiveness of those interventions over time. Frequently reported chains can help physicians and public health experts understand the correlations and causal relationships between clinical diseases, potentially allowing the discovery of causal relationships that had not been previously observed. On the patient-level, personal medicine perspective, it may even be possible to warn individual patients of potential diseases leading to death even before any symptoms can be diagnosed. This can help improve clinical care and, in turn, the well-being of patients.

To facilitate the timely, accurate and complete reporting of deaths and reduce the subjectivity of reporting physicians, in this paper, we aim to develop a decision support system that suggests probable causal sequences of death based on decedents’ health histories. These chains of causes of death form the basis of the NCHS Multiple Causes of Death data, a critically valuable data source in public health. These chains outline the chain of medical events and conditions which led to the death, arranged in a cause-effect order. 
We identify three challenges in this task with the obtained mortality data set when predicting the causal chain of death using deep learning algorithms. \textbf{Table \ref{tab:challengesAndSolutions}} summarizes these three challenges and our proposed solutions.

\begin{table}[h!]
    \centering
    \caption{Summary of challenges in predicting causal chain of death and proposed solutions.}
    \begin{tabular}{|m{3.5cm}|m{4cm}|}
    \hline
    Challenge & Solution \\
    \hline
    Different coding versions & Machine translation between input and output sequences \\
    \hline
    Domain knowledge Conflict & Incorporate medical domain knowledge as constraint \\
    \hline
    Data interoperability & FHIR compatible platform \\
    \hline
    \end{tabular}
    \label{tab:challengesAndSolutions}
\end{table}

The first challenge comes from the different coding systems of clinical conditions. The existing causes of death (COD) in the United States have been using the International Statistical Classification of Diseases and Related Health Problems, Tenth Revision (ICD-10) since January 1999 \footnote{https://www.cdc.gov/nchs/icd/icd10cm.htm} \cite{world2004international}. On the other hand, healthcare institutions and practitioners in the U.S. were still filing patients' health record using the ninth revision (ICD-9) codes until October 2015 \cite{khera2018transition}. ICD-10 codes are "very  different" from ICD-9 codes in both coding structure and quantity: ICD-10 has nearly five times as many diagnosis codes as ICD-9 \footnote{https://www.cdc.gov/nchs/icd/icd10cm\_pcs\_faq.htm}.

Intuitively, a solution to this challenge rises from the analogy between our task and the natural language translation. The input sequence of diagnosis codes is from the last hospital discharge record of the deceased, and the output sequence is the corresponding causes of death for that decedent. Similar to translating from English sentences to French sentences, we intend to propose a succinct causal chain of death in ICD-10 codes from the priority-based discharge records of ICD-9 codes. The research area of Natural Language Processing (NLP) contains extensive studies for machine translation. Specifically, the models for machine translation can be classified into autoregressive (AR) \cite{bahdanau2014neural}\cite{dai2015semi}\cite{luong2015effective}\cite{luong2016achieving} and autoencoder (AE) models \cite{Devlin2018bert}\cite{yang2019xlnet}\cite{lample2019cross}, with the former factorizing the probability of a given corpus into a series of conditional probabilities and the latter generating output through reconstructing from corrupted input.

The second challenge is the domain knowledge conflict. As a pure data-driven approach, the deep learning model can sometimes generate confusing sequences to the physicians. Some results may even contradict the medical domain knowledge. Consequently, the physicians may find it difficult to trust the generated results. To solve this problem, we would like to incorporate medical domain knowledge to guide the deep learning framework. Particularly, we use an external source of expert-curated rules that are pairs of causal relationships between clinical condition codes. When the deep learning model searches for the next clinical condition in the process of generating output sequences, we impose the constraint that only clinical conditions following medical domain knowledge can serve as candidates.

The last challenge is the data interoperability in death reporting. Currently, the National Center for Health Statistics (NCHS) coordinates with 57 reporting jurisdictions across the United States to aggregate mortality data \cite{cowper2002primer}. These reporting jurisdictions have different regulations and local laws. To streamline the data storage and transmission between hospitals and these public health institutions and to increase data size for future big data analytics, Fast Healthcare Interoperability Resources (FHIR) is utilized to standardize mortality data reporting. We have developed one web-based FHIR \cite{hoffman2018intelligent} platform that adopts the new health standard named HL7 \cite{dolin2006hl7} to access electronic health records data. The newly developed Android version mobile app is FHIR compatible; it is capable of pre-populating different sections of death certificate to extract essential information of health history of the decedents. Furthermore, it serves as a graphical user interface for physicians that the mobile app can automatically query the pre-trained causal chain of death prediction models to provide decision support. In future versions, such applications may collect data that can be used to refine and train decision support models in real-time, expanding the impact of this work beyond retrospective analysis and improving clinical practice at the point-of-care.

\begin{figure*}[t!]
    \centering
    \includegraphics[scale=0.32]{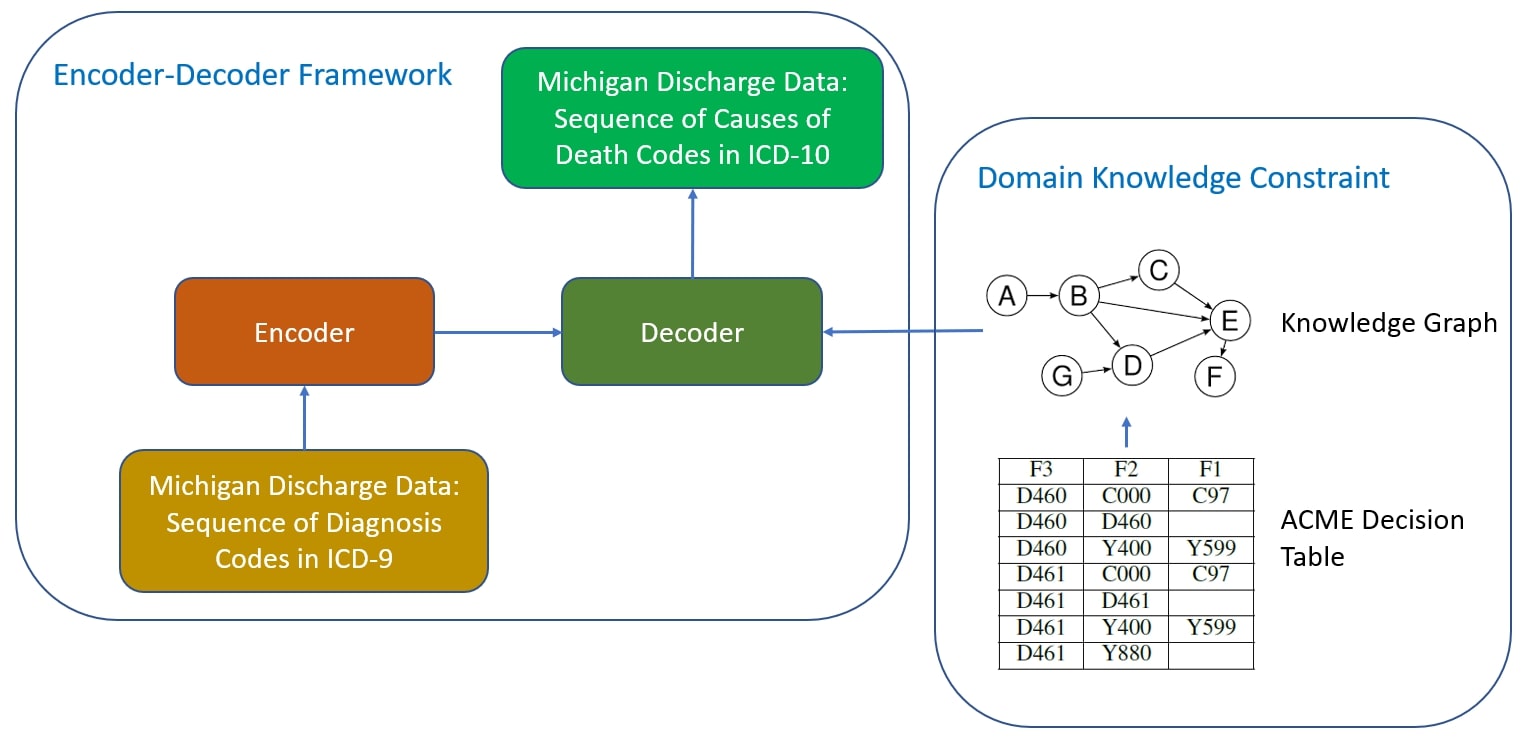}
    \caption{Overall Structure of this paper. The encoder-decoder model is the main framework for generating causal chain of death. Domain knowledge constraint is learnt from ACME decision table.}
    \label{fig:overall_structure}
\end{figure*}

In this work, we identify the encoder-decoder models as the main framework to automatically generate a causal sequence of death in ICD-10 given priority-based diagnosis codes in ICD-9 from one decedent's last hospital discharge record. Meanwhile, we demonstrate the feasibility of the encoder-decoder models for ICD-10 input data by mapping ICD-9 codes to ICD-10 codes as preprocessing. This makes our work applicable to ongoing clinical practice, as we are using ICD-10 codes in current electronic health records data. In addition, we also add expert domain knowledge graph learnt from ACME decision table as knowledge constraint to restrict the output from the pure data-driven framework. The overall structure is shown in \textbf{Figure \ref{fig:overall_structure}}.

In summary, our work has the following contributions: 
\begin{enumerate}
    \item We are the first to develop encoder-decoder data-driven approaches for suggesting causal chains of death based on death reports and decedents' last hospital visit discharge records;
    \item We apply the State-of-The-Art model for neural machine translation, and augment it with domain knowledge constraints;
    \item We demonstrate the effectiveness of validity check in improving the stability of encoder-decoder frameworks;
    \item We are the first one to interpret deep learning results through visualization by identifying meaningful association between clinical conditions coded in different versions;
    \item We are the first one to use BLEU score to evaluate the performance of deep learning model on generating causal chain of death;
    \item We implement the knowledge-guided deep model on the FHIR interface.
\end{enumerate}
\section{Recent Work}

Intelligent death reporting has been a rising research field in recent years. Jiang et al. \cite{jiang2017causes} applied topic modeling on the multiple causes-of-death US mortality data from NCHS between 1999 and 2014. The author successfully group morbidities based on their correlation and explore the temporal evolution of these morbidity groups. Yet there is one major limitation with this work: due to the nature of unsupervised learning, the author failed to determine the optimal number of topic groups, negating its impact on clinical practice. Wu and Wang \cite{wu2017infer} designed a convolutional neural network (CNN) with dynamic computation graph to infer the underlying cause of death using the same NCHS mortality data. Using a list of relevant medical conditions, the proposed CNN model was able to achieve 75\% accuracy in predicting the single underlying cause of death. Besides the high accuracy, the proposed method further assist physicians in death reporting by providing human understandable interpretation for the model output. Meanwhile, Hoffman et al. \cite{hoffman2018validity} revealed the poor quality of death reporting data by showing 20.1\% discordance of cause of death. The author also proposed validity check on death reporting data to remove invalid causal pairs of death codes. One limitation is that, the author didn't apply validity check on any downstream tasks, such as predicting the single underlying cause of death, to further demonstrate the value of validity check. 

\begin{figure*}[t!]
    \centering
    \includegraphics[scale=0.45]{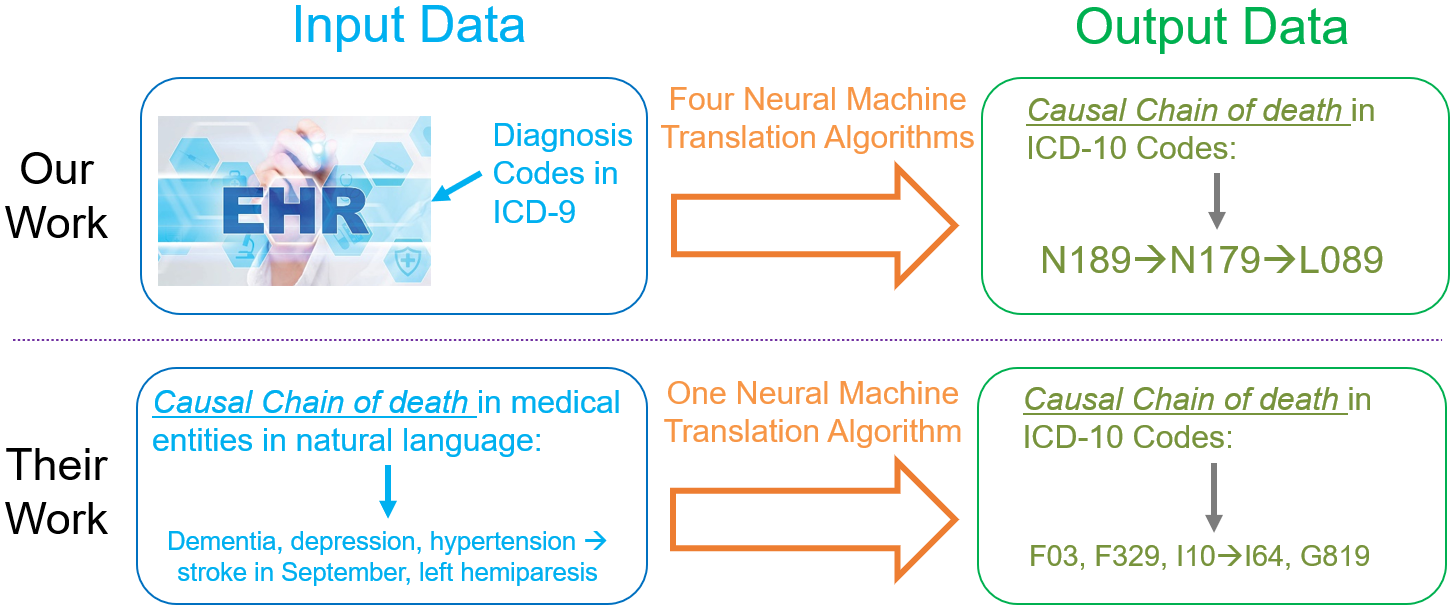}
    \caption{The top row show the pipeline of our work and the bottom row show the pipeline of Falissard's work.}
    \label{fig:pipeline}
\end{figure*}

A recent work of similar topic \cite{falissard2020neural} was submitted in late March, 2020 and \textbf{Figure \ref{fig:pipeline}} shows the difference in pipeline of our work and their work. Our work has five key differences from theirs: 
\begin{enumerate}
    \item Different clinical tasks: we aim to automatically generate the causal chain of death in ICD-10 codes given the discharge data of the decedent's last hospital visit, coded in ICD-9. In comparison, their work is to recognize and convert medical entities in natural language (French) to ICD-10 codes. The medical entities in French are already filled as causes of death on the death certificate by medical experts.
    \item Different input data formats: our input data are priority-based ICD-9 codes from the discharge data of the decedent's last hospital visit, while their input data are medical entities of causes of death in French on the death certificates. 
    \item Different methods: in addition to neural machine translation model named transformer, we also adapt and apply three recurrent neural network based encode-decoder frameworks to generate the casual chain of death; furthermore, we test the more recent the cross-lingual language modeling (XLM) \cite{lample2019cross} and analyze why it fails on our task.
    \item Medical domain knowledge constraint: we learn medical domain knowledge as constraint from ACME decision table on the encoder-decoder frameworks.
    \item Different evaluation metrics: we use the modified BLEU score (1-gram and 2-gram precision), which is popular for sequence-to-sequence translation task in natural language processing. Their work uses precision, recall and f-measure for evaluation. 
\end{enumerate}
\section{Causal Chain of Death}

\subsection{Data}
\begin{figure*}[!]
    \centering
    \includegraphics[width=\textwidth]{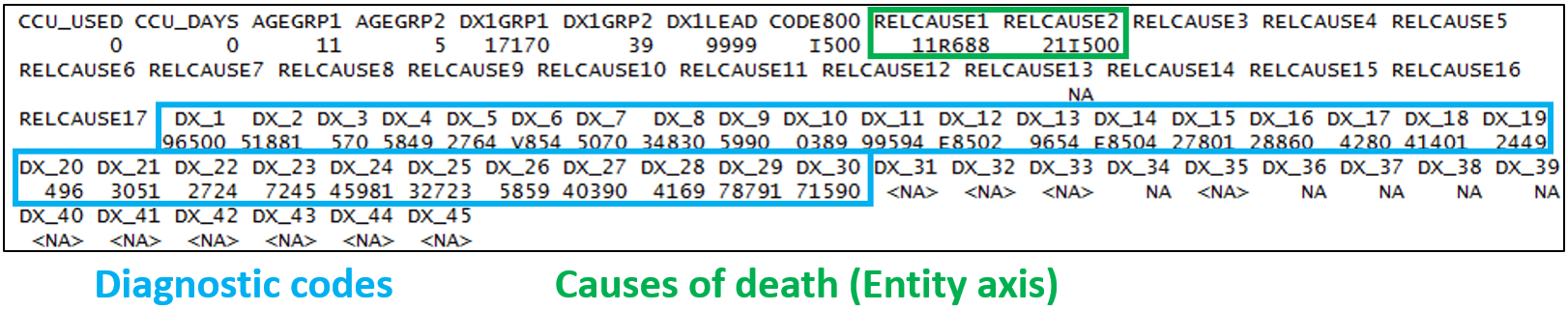}
    \caption{Sample data from the Michigan data set. The casual chain of death in ICD-10 for this patient is $I500>R688$ (Heart failure $>$ Other general symptoms and signs), outlined in green. This patient had a total of $30$ ICD-9 diagnostic codes assigned during the last visit to hospital, outlined in blue.}
    \label{fig:michigan-data-screenshot}
\end{figure*}

We are using last hospital visit discharge records from Michigan Vital Statistics Data that covers $181,137$ patients. As shown in \textbf{Figure. \ref{fig:michigan-data-screenshot}}, each patient has exactly one line of last hospital visit essential information, including up to $45$ clinical diagnosis codes, one underlying cause of death and up to $17$ related causes of death. On average, each patient has $18.84$ diagnosis codes and $2.25$ causes of death (including the underlying cause of death). In line with the ICD-9-CM Official Guidelines for Coding and Reporting \footnote{https://www.cdc.gov/nchs/data/icd/icd9cm\_guidelines\_2011.pdf}, the diagnosis codes are in priority-based sequence of ICD-9 codes. The causes of death are in ICD-10 codes. Typically, we shall have a longer input source sequence around 16-20 codes, and a much shorter output target sequence with roughly 2-3 codes. Such a short sequence of death codes is expected in death reports. We accessed the ten years’ (2009 to 2018) National Center for Health Statistics (NCHS) Mortality Multiple Cause Files database \footnote{https://www.cdc.gov/nchs/data\_access/vitalstatsonline.htm\#Mortality\_Multiple} and calculated that the average length of death code sequence among $26,322,220$ decedent samples to be 2.95 codes. (Note that discharge codes on last hospital admission may contain previous admission discharge codes.)

Ontology of medically valid causal relationships between ICD-10 codes were developed, improved, and promulgated by an international team of medical experts \cite{lu2003using}. This ACME (Automatic Classification of Medical Entry) decision table was used to learn the medical domain knowledge constraint \cite{hoffman2018validity}. It contains 95,321 lines of causal relationship. \textbf{Table \ref{tab:ontology}} illustrates an example of this decision table. Specifically, if rules are of length 2 it can be interpreted as F2 $\rightarrow$ F3, underlying cause of death leading to immediate cause of death; if rules are of length 3, it can be represented as (F1:F2) $\rightarrow$ F3, a subset of underlying cause of death leading to the immediate cause of death.  

\begin{table}[h!]
    \centering
    \caption{An Example of ACME Decision Table.If rules are of length 2 it can be interpreted as F2 $\rightarrow$ F3, underlying cause of death leading to immediate cause of death; if rules are of length 3, it can be represented as (F1:F2) $\rightarrow$ F3, a subset of underlying cause of death leading to the immediate cause of death.}
    \begin{tabular}{|c|c|c|}
        \hline
        F3 & F2 & F1 \\
        \hline
        D460 & C000 & C97 \\
        \hline
        D460 & D460 & \\
        \hline
        D460 & Y400 & Y599 \\
        \hline
        D461 & C000 & C97\\
        \hline
        D461 & D461 & \\
        \hline
        D461 & Y400 & Y599 \\
        \hline
        D461 & Y880 & \\
        \hline
    \end{tabular}
    \label{tab:ontology}
\end{table}

\subsection{Generating Causal Chains Through Translation}
Mathematically, we can define the generation of causal chains as follows:

\theoremstyle{definition}
\begin{definition}{[Generation of Causal Chains]}
Given a deceased's medical history represented as a collection of medical codes $\mathbf{x} = x_1, \dots, x_m$, the goal of causal chain generation is to identify another list of medical codes $\mathbf{y} = y_1, \dots, y_n$ that summarizes the conditions leading to the death.
\end{definition}

The objective is to propose the causal chain of death, which is an ordered sequence of death codes (ICD-10). The inputs, on the other hand, are sequences of diagnosis codes (ICD-9). To generate one sequence of one domain (language) from a sequence of another domain (language), we apply the state-of-the-art algorithms from neurall machine translation.

Diagnosis codes in ICD-9 are defined as source codes (input sequence) while death codes in ICD-10 are target codes (output sequence). Both source codes and target codes are split into training set, validation set and testing set according to the ratio of $7:1:2$. We also applied 5-fold cross validation. Each line of codes is treated as one sequence, with each code in the sequence treated as one word in the natural language settings.
\section{Methodology}

\subsection{Neural Machine Translation: Encoder and Decoder}

\begin{figure}[h!]
    \centering
    \includegraphics[scale=0.25]{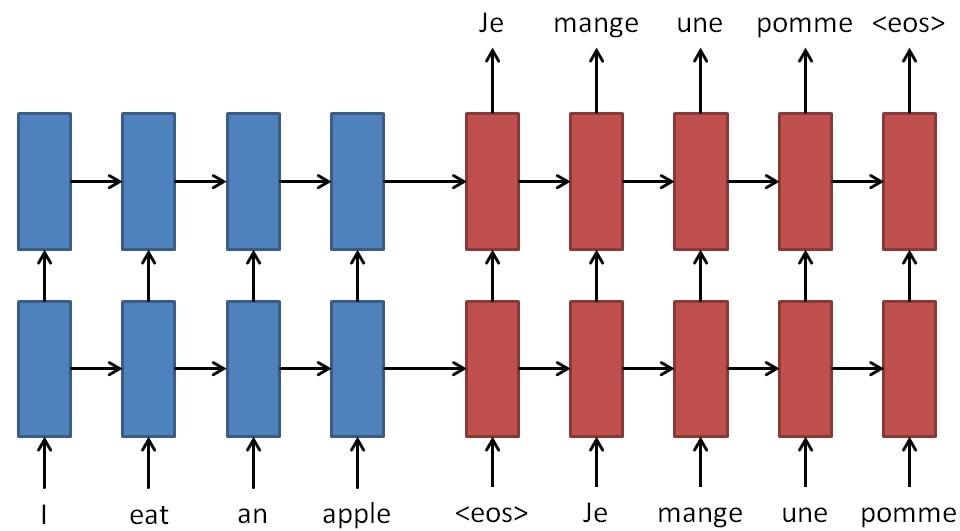}
    \caption{Neural machine translation consists of an encoder (stacked recurrent networks in blue) and a decoder (stacked recurrent networks in red). The symbol $<eos>$ is a special token referring to the end of a sentence. Adapted from \cite{luong2016achieving}.}
    \label{fig:NMT}
\end{figure}

Translation is to find a target sentence $\mathbf{y} = y_1, \dots, y_n$ which maximizes the conditional probability $p(\mathbf{y}|\mathbf{x})$ given a source sentence $\mathbf{x} = x_1, \dots, x_m$. Neural machine translation (NMT) aims to maximize this conditional probability of source-target sentence pairs by using a parallel training corpus to fit a parameterized model. As shown in \textbf{Figure. \ref{fig:NMT}}, there are two basic components of an NMT system: 
\begin{itemize}
    \item An encoder encodes the input sequence $\mathbf{x}$ into a representation $\mathbf{s}$
    \item A decoder generates the output sequence $\mathbf{y}$
\end{itemize}

The conditional probability of the decoder is formulated as: 
\begin{equation}
    \log p(\mathbf{y}|\mathbf{x}) = \sum_{t=1}^n \log p(y_t|y_1, y_2, \dots, y_{t-1}, \mathbf{s})
\end{equation}

The the probability of the next generated word $y_i$, is jointly decided by the learned representation $\mathbf{s}$ and all previously generated words $y_1, \dots ,y_{t-1}$.

In this section, we would like to briefly introduce the following four models of encoder-decoder framework in neural machine translation.

\subsubsection{LSTM Encoder - LSTM Decoder}
In an long short-term memory (LSTM) Encoder-Decoder framework \cite{cho2014learning,sutskever2014sequence}, the encoder reads and encodes an input sequence of embedded vectors $\mathbf{x}$. The encoder will then generate a hidden state $h_t$ at time $t$ from the current input $x_t$ and the previous hidden state $h_{t-1}$:
\begin{equation}
    h_t = f(x_t, h_{t-1})
\end{equation}

The representation vector $\mathbf{s}$ shall have the form:
\begin{equation}
    \mathbf{s} = q({h_1, ..., h_m})
\end{equation}

Here $f$ and $q$ are some non-linear functions. For the basic recurrent neural network RNN/LSTM model, the conditional probability of output sequence $\mathbf{y}$ at time $t$ can be written as: 
\begin{equation}
    p(y_t|y_1, ..., y_{t-1}, \mathbf{s}) = g(y_{t-1}, h_t, \mathbf{s})
\end{equation}

Here $g$ is a (multi-layered) nonlinear function.

Generic RNN or LSTM encoder-decoder framework has to process the sentence word by word, failing to preserve long-term dependency. Luong et al \cite{luong2015effective} proposed global attention which predicts the position of alignment for the current word before computing the context vector using the window centered around that source position. Here the global attention is used in the decoder. The LSTM encoder-decoder model is easy to understand, and can be applied on most sequence-to-sequence tasks. Yet such a model has limited performance, especially on long sentences.

\subsubsection{Mean Encoder - LSTM Decoder}

Mean encoder is a simplified encoder. Instead of adding an LSTM model in the encoder, this mean encoder speeds up by using a mean pooling layer. The same stacked LSTM model and global attention mechanism are used in decoder. This mean-encoder model is much more efficient during training than any other models, but may not accurately capture the long-term dependency in the input sentences.

\subsubsection{Bidirectional RNN Encoder - LSTM Decoder}

A major disadvantage of the traditional encoder-decoder model is that the neural networks compress source sentences into fixed-length vectors. This may significantly limit the capability of translating long sentences \cite{cho2014properties}. Bahdanau proposed to use a bidirectional RNN \cite{bahdanau2014neural} with encoder-decoder approach so that the model can learn to align and translate jointly. A bi-directional RNN encoder model can better learn the embedding of words, but it's less efficient than the LSTM encoder-decoder framework, and typically has less accurate results than transformer models.



\begin{figure}[h!]
    \centering
    \includegraphics[scale=0.25]{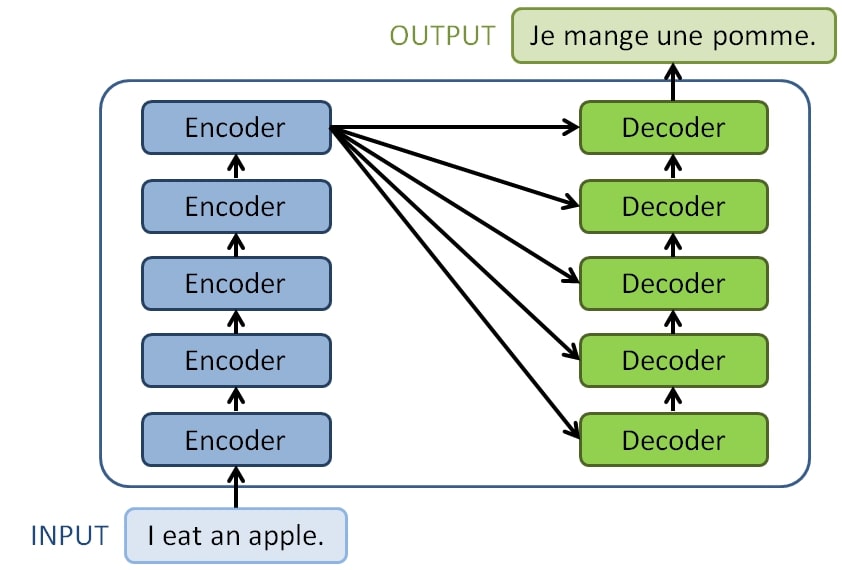}
    \caption{Overall structure of a transformer. Here we have five identical encoders and five identical decoders in this transformer.}
    \label{fig:transformer}
\end{figure}

\subsubsection{Transformer Model}
Still, neither RNN-based or CNN-based encoder-decoder models perform well on long sentences. To overcome this problem, Vaswani et al proposed the transformer framework \cite{vaswani2017attention} that enables encoding words of the same sentence in parallel. Together with self-attention, transformer boosts the speed of encoding. 

As shown in \textbf{Figure. \ref{fig:transformer}} \footnotemark
, a transformer consists a stack of encoders and the same number of decoders. The embedded input is passed to the encoder at the bottom; the output from the encoder on the top will be passed to all decoders. The decoder on the top will pass the output to a linear layer and a softmax layer to generate predicted sentence.

\begin{figure}[h!]
    \centering
    \includegraphics[scale=0.22]{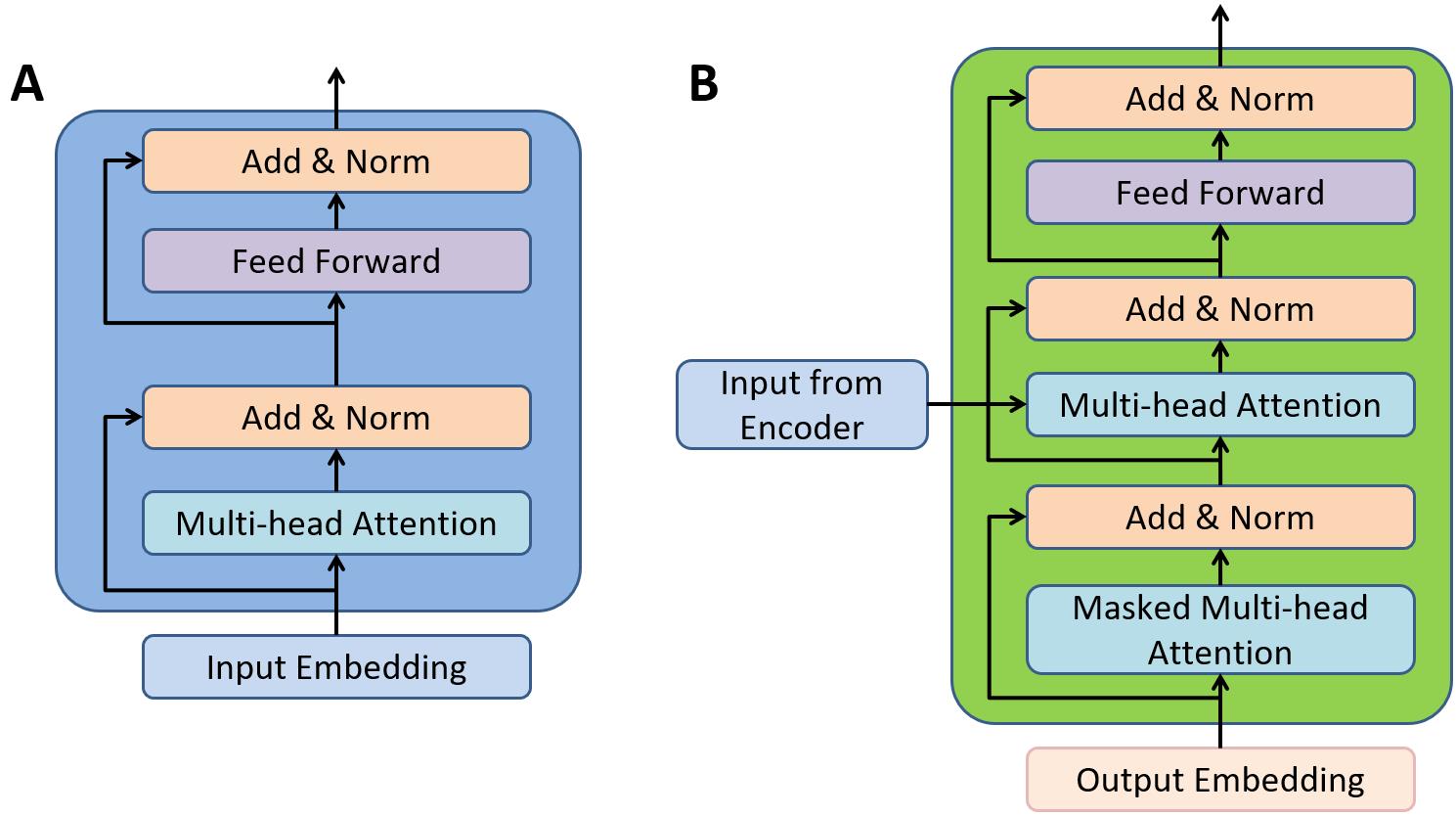}
    \caption{Detailed structure of a transformer. \textbf{A} is the encoder of the transformer. It consists of two layers: the feed forward layer and multi-head attention layer. \textbf{B} is the decoder of the transformer. It consists of three layers: the feed forward layer and two multi-head attention layers.}
    \label{fig:encoder-decoder-transformer}
\end{figure}

Encoder has two layers: a multi-head self-attention layer and a feed forward layer (shown in part A of \textbf{Figure. \ref{fig:encoder-decoder-transformer}} \footnotemark[\value{footnote}]). Decoder has an extra multi-head attention layer that processes both the output from the encoder stack and the output from previous multi-head attention layer (shown in part B of \textbf{Figure. \ref{fig:encoder-decoder-transformer}} \footnotemark[\value{footnote}]).

The transformer model is more time-consuming than RNN-based encoder-decoder frameworks during the training stage, but can achieve far better results \cite{vaswani2017attention}. BERT (Bidirectional Encoder Representations from Transformers) \cite{Devlin2018bert} is a transformer encoder model that has been pre-trained on large datasets (BooksCorpus with 800M words and English Wikipedia with 2,500M words). The pre-trained BERT model can be further fine-tuned to improve performance on multiple nature language processing tasks.

\footnotetext{Adapted from https://towardsdatascience.com/transformers-141e32e69591}


\subsection{Decoding and Translation}

A straightforward method of decoding is to keep and predict only one word with the highest score based on previous steps. It is efficient and easy to understand; yet a small mistaken output might corrupt all remaining predictions. Thus, a better strategy named beam search keeps the top $k$ hypotheses for each step and select the best one when reaching the end of sequence. Here, $k$ is the beam size. 

We also include medical domain knowledge as constraints during translation. The ACME (Automatic Classification of Medical Entry) decision table specifies all the ``feasible'' pairwise causal relationships between ICD diagnosis codes \cite{hoffman2018validity,lu2003using}. Using this decision table, we construct a domain knowledge graph on all diagnosis codes from Michigan data before training. 
With diagnosis codes as nodes, we add directed paths between them only if such causal relationship can be found in the ACME decision table. 
When decoding, the networks are required to look up the knowledge graph and are only allowed to include "feasible" codes in the top $k$ hypotheses.

\subsection{Evaluation}
Furthermore, we need to increase the interpretability and demonstrate it through visualization. Several studies visualized attention weights to showcase that their models captured the correspondence words between English and French \cite{bahdanau2014neural} \cite{yang2016hierarchical} \cite{vaswani2017attention}. The ability to demonstrate the correspondence of words between languages through visualization is also critical for our task. Because the discharge records and causes of death are coded under different versions of ICD codes, associating a cause of death to several past diseases could help us qualitatively evaluate generated causal sequences.

In addition to qualitative evaluation, quantitative evaluation remains critical. Here we evaluate how well our proposed causal chain $\hat{Y}=\{\hat{Y}_1, .., \hat{Y}_{M_1}\}$ aligns with the physicians' decision, i.e., $Y = \{Y_1, .., Y_{M_2} \}$.
Here $Y_i$ is the individual codes, and $M_1, M_2$ are the respective length of the chains.
A perfect alignment means $M_1 = M_2$, and $\hat{Y}_i = Y_i$, for $i=1, ..., M_1$.
However, this is rarely the case, thus we compute a weighted average precision of our alignment in sub-sequences of variable lengths, i.e., the BLEU (BiLingual Evaluation Understudy) score \cite{papineni2002bleu}. Following natural language processing literature, we call sub-sequence of length $i$ "i-grams".
BLEU score ranges from 0 to 1 or (or from 0 to 100 if multiplied by 100), and the higher BLEU, the higher we have an alignment with physicians clinically.

We use a simple example as follows to illustrate the computation of the BLEU score. In our proposed candidate sequence, the underlying cause of death, \textit{Asphyxia and Hypoxemia (R909)} leads to \textit{Pneumonia, Unspecified Organism (J189)} which leads to \textit{Respiratory failure, unspecified (J969)}.

$$\hat{Y} = R909 \rightarrow J189 \rightarrow J969 $$

The reference sequence, determined by the physician, consists of \textit{Asphyxia and Hypoxemia (R909)}, \textit{Pneumonia, Unspecified Organism (J189)} and then \textit{Acute Respiratory Failure (J960)}.

$$ Y = R909 \rightarrow J189 \rightarrow J960 $$

As shown in \textbf{Table \ref{tab:bleu}}, we first list 1-grams and 2-grams from $\hat{Y}$ and $Y$, and we compute the precision for the two case. Here the definition of precision is similar in the classification setting: among all the predictions we made in candidate sequence $\hat{Y}$, how many we get correct in the reference sequence $Y$?
After we compute all the precision metrics, we calculate the geometric average of them as the BLEU metrics, in this case, approximately 0.47.

\begin{table}[!h]
\centering
\caption{An example of 1-gram precision and 2-gram precision in BLEU score.}
\label{tab:bleu}
\resizebox{.49\textwidth}{!}{%
\begin{tabular}{|l|l|l|r|}
\hline
Grams & From Candidate $\hat{Y}$ & Appear in $Y$ & Precision \\ \hline
1-gram & (R909), (J189), (J969) & (R909), (J189) & 2/3 \\ \hline
2-gram & (R909, J189), (J189, J969) & (R909, J189) & 1/2 \\ \hline
\end{tabular}%
}
\end{table}

In natural language settings, people usually calculate BLEU score for the geometric average up to 4-gram precision. In our case, however, we only compute the geometric average up to 2-gram precision, and apply clipping to each of the precision. This is due to the fact that the average length of causal chain of death in Michigan dataset is $2.25$ codes so including 3-gram precision will lead to substantially inaccurate evaluation. Furthermore, we also include a brevity penalty to penalize sentences that are too short. 
$$BLEU = min\{ 1, \exp{(1-\frac{length(Y)}{length(\hat{Y})})} \} (\prod_{i=1}^2 precision_i)^{1/2} $$
where $precision_i$ is the i-gram precision, defined as
$$\frac{\sum_{\hat{Y}} \text{the count of i-grams in } \hat{Y} \text{ that appears in } Y}{ \sum_{\hat{Y}} \text{the count of i-grams in } \hat{Y}}$$

In this equation, $exp$ is the natural exponential function; $length(Y)$ is the length of reference sequence (how many words in the reference sequence); likewise, $length(\hat{Y})$ is the length of the proposed sequence (how many words in the proposed sequence). We refer readers to the original paper for a detailed discussion on the variants of BLEU metric \cite{papineni2002bleu}.

For clinical interpretation, our modified BLEU score indicates how well our proposed sub-sequences of causal conditions match the physicians' results. The 1-gram precision emphasizes individual condition codes matching, while 2-gram precision evaluates the causal relationship between two neighboring condition codes. Physicians can manually check whether the generated causal relationship between any two neighboring condition codes fulfills or contradicts their medical domain knowledge; in addition, a data-driven algorithm can incorporate ACME decision table as medical domain ground truth to assess the validity of two neighboring condition codes. 

In \textbf{Table \ref{tab:bleu_example_2}}, we show an example of different candidate sequences that have perfect 1-gram precision but different 2-gram precision. The reference sequence from underlying cause of death to immediate cause of death is: \textit{I251 (Atherosclerotic heart disease of native coronary artery), I38 (Endocarditis, valve unspecified), I429 (Cardiomyopathy, unspecified) and I469 (Cardiac arrest, cause unspecified)}. We argue that our modified BLEU score favors candidate sequences that have more reasonable and feasible condition codes with pairwise casual relationship.

\begin{table}[h!]
    \centering
    \caption{Our modified BLEU score for different candidate sequences.}
    \begin{tabular}{|l|l|r|}
    \hline
         & Sequence & BLEU \\ \hline
        Reference & $\textcolor{ForestGreen}{I251} \rightarrow \textcolor{Dandelion}{I38} \rightarrow \textcolor{Magenta}{I429} \rightarrow \textcolor{Red}{I469}$ & \\ \hline \hline
        Candidate 1 & $\textcolor{Magenta}{I429} \rightarrow \textcolor{Dandelion}{I38} \rightarrow \textcolor{Red}{I469} \rightarrow \textcolor{ForestGreen}{I251}$ & 0.0 \\ \hline
        Candidate 2 & $\textcolor{Dandelion}{I38} \rightarrow \textcolor{Magenta}{I429} \rightarrow \textcolor{ForestGreen}{I251} \rightarrow \textcolor{Red}{I469}$ & 57.7 \\ \hline
        Candidate 3 & $\textcolor{Magenta}{I429} \rightarrow \textcolor{Red}{I469} \rightarrow \textcolor{ForestGreen}{I251} \rightarrow \textcolor{Dandelion}{I38}$ & 81.6 \\ \hline
        Candidate 4 & $\textcolor{Dandelion}{I38} \rightarrow \textcolor{Magenta}{I429} \rightarrow \textcolor{Red}{I469} \rightarrow \textcolor{ForestGreen}{I251}$ & 81.6 \\ \hline
        Candidate 5 & $\textcolor{ForestGreen}{I251} \rightarrow \textcolor{Dandelion}{I38} \rightarrow \textcolor{Magenta}{I429} \rightarrow \textcolor{Red}{I469}$ & 100.0 \\ \hline
    \end{tabular}
    \label{tab:bleu_example_2}
\end{table}


In addition to our modified BLEU score, we also include three other evaluation criteria: the accuracy for predicting the entire output sequence correctly, the accuracy of predicting single code correctly in the output sequence, and the accuracy for predicting the underlying cause of death correctly.
\section{Experiments}

By using OpenNMT package \cite{opennmt}, we have tested five different encoder-decoder models. In addition to OpenNMT, we have incorporated the state-of-the-art pretraining model named cross-lingual language model (XLM) \cite{lample2019cross} on our data set.

To extend the scope of this work, we would like to explore the feasibility of applying encoder-decoder frameworks on current electronic health records (EHRs) data in ICD-10 codes. As the input data from the Michigan dataset is coded in ICD-9, we choose to map the input ICD-9 codes into ICD-10 codes using General Equivalence Mappings published by Centers for Medicare \& Medicaid Services (CMS) \footnote{https://www.nber.org/research/data/icd-9-cm-and-icd-10-cm-and-icd-10-pcs-crosswalk-or-general-equivalence-mappings}.

Overall, we have conducted four experiments on ICD-9 input codes (four combinations with or without validity check, with or without knowledge constraint) and one experiment on ICD-10 input codes without validity check and knowledge constraint.

\subsection{OpenNMT}

OpenNMT serializes the training, validation and vocabulary data into PyTorch files for preprocessing. As the Michigan data have a relatively small sample size comparing with datasets used in natural language processing, our models have a small number of parameters but similar architecture as the state-of-the-art models. During training, we use the 2-layer LSTM model, with 500 hidden units in each layer for the default LSTM encoder framework (Luong et al. used 4-layer LSTM model with 1000 units \cite{luong2015effective}). The mean encoder framework simply has the two LSTM layers removed in encoder but keeps other parts unchanged. For bidirectional RNN encoder, a 2-layer bidirectional LSTM with 500 and 250 hidden units is implemented. CNN-based encoder includes two gated convolutional layers with 500 and 1000 hidden units and kernel size (3,1) followed with two convolutional multi-step attention layers. The transformer has 6 stacking layers, with $2048$ hidden units in feed forward layers and 8 heads in multi-head attention layers.

We used one Nvidia GPU Tesla K80 to train and evaluate the models. Typically it took around 2-6 hours to train an LSTM, CNN or bidirectional RNN model, and about 18 hours to train a transformer model. Yet it took less than five minutes to translate all 36,000 testing data using any of these models.

\subsection{XLM: Pretraining}

XLM (cross-lingual language model) \cite{lample2019cross} incorporates masked language modeling (MLM) proposed in BERT (Bidirectional Encoder Representations from Transformers) \cite{Devlin2018bert} with the transformer model to improve translation performance. The preprocessing includes tokenizing and applying fastBPE (byte pair encoding) \cite{sennrich2015neural} to monolingual and parallel data.  MLM is the core strategy in monolingual language model pretraining. Training consists of three major steps: denosing auto-encoder, parallel data training and online back-translation.

Due to the limited size of our data set, we concatenate all training, validation and testing data into two corpora for monolingual pre-training. Masked language modeling (MLM) perplexities are used for validation during pre-training. We further train the cross-lingual model with parallel validation data and predict on parallel test data.

We set the transformer framework with 512 embedding size and 4 attention heads. We vary the encoder-decoder stacking size from 6 layers to 1 layer. The drop out rate is 0.1, attention dropout 0.1, batch size 32 and sequence length 128. We used GELU for activation and adam as optimizer. 

\subsection{Optional Preprocessing: Validity Check}

In search for better prediction performance, we add an extra pre-processing step, the validity check. For training and validation data, we adopt the same algorithm in \cite{hoffman2018validity} to remove the pairs of sentences that include "invalid" causal relationship between diagnosis codes in target sentence. In this way we reduce the number of sentences in the training set from $136,753$ to $107,711$ and those in the validation set from $34,385$ to $27,009$. We then follow the same pipeline to train and translate with the same five models. 
\section{Results and Discussion}
\subsection{OpenNMT}


\begin{table*}[!]
    \centering
    \caption{Average BLEU scores and standard deviation in parentheses across five folds.}
    \begin{tabular}{|c|l|r|r|c|c|c|c|}
         \hline
         Experiment & Input Data & Validity Check & Knowledge Constraint & LSTM & Mean & BRNN & Transformer \\
         \hline
         1 & ICD-9 & Not checked & Non-constrained & 15.46 (0.71) & 14.81 (0.48) & \textbf{16.04} (0.74) & 14.37 (0.55) \\
         \hline
         2 & ICD-9 & Checked & Non-constrained & 15.58 (0.48) & 14.11 (0.34) & 15.95 (0.44) & 14.38 (0.30) \\
         \hline
         \hline
         3 & ICD-9 & Not checked & Constrained & 12.61 (6.33) & 15.56 (0.72) & 12.46 (6.26) & 14.76 (0.51) \\
         \hline
         4 & ICD-9 & Checked & Constrained & 12.95 (6.50) & 15.15 (0.48) & 13.16 (6.61) & 14.99 (0.42) \\
         \hline
         \hline
         5 & ICD-10 & Not checked & Non-constrained & 15.47 (0.71) & 14.93 (0.46) & 15.72 (0.66) & 14.40 (0.63) \\
         \hline
    \end{tabular}
    \label{tab:openNMT-results-table}
\end{table*}

As shown in \textbf{Table \ref{tab:openNMT-results-table}}, we calculate the average BLEU score and its standard deviation in parenthesis for each encoder-decoder framework across five folds. For \textbf{Experiment 1} which is without validation check in training data and knowledge constraint in decoding, the bi-directional RNN encoder-decoder model achieved the highest average BLEU score, followed by the LSTM encoder-decoder model, the mean encoder-decoder model and the transformer model. Comparing \textbf{Experiment 1} and \textbf{Experiment 2}, it is clear that validity check, the preprocessing step on training and validation data, can help reduce the standard deviation of the BLEU score. The average BLEU score between \textbf{Experiment 1} and \textbf{Experiment 2}, however, is very close for LSTM (difference=-0.12), BRNN (difference=0.09) and transformer (difference=-0.01) models. This indicates that validity check can help improve stability of the trained models but cannot improve the average performance.

It is worth noticing that in \textbf{Experiment 3} and \textbf{Experiment 4}, the average BLEU score drops significantly for LSTM and BRNN models, while their standard deviation increases significantly. As for the mean encoder and transformer models, knowledge constraint slightly increases both the average BLEU score and its standard deviation. In other words, applying knowledge constraint has a mixed impact on the encoder-decoder models: it may significantly decrease the performance and stability on LSTM and BRNN models, while slightly improve the performance on the other two models. Consequently, we argue that the encoder-decoder frameworks can learn the causal relationship between diagnosis codes well enough that it is not necessary to learn and incorporate the medical domain knowledge constraint from the ACME decision table 

In addition, it is interesting to compare the results in \textbf{Experiment 1} and \textbf{Experiment 5}. After mapping the input ICD-9 codes into ICD-10 codes, LSTM, mean encoder and transformer models have similar average BLEU scores with those in Experiment 1. The average BLEU score of BRNN model drops by 0.32, still relatively small (less than 2\% difference). The standard deviation for all models are very close between two experiments. These results are significant: 1) the encoder-decoder frameworks are promising and stable in generating the causal chain of death, no matter whether we have input and output data in the same or different coding systems. 2) When having no access or limited access to the newest electronic health records (EHRs) data, we can use data before 2015 to train the models and generate the causal chain of death.

According to \cite{luong2016achieving}, larger vocabulary size tends to achieve higher BLEU score. Their proposed hybrid NMT model achieved $17.7$ BLEU score with $10k$ vocabulary size on English-Czech translation task. Our vocabulary size in source set is $7616$ and that in target set is $2649$. Thus, our translation performance is close to the state-of-the-art.

\begin{table*}[!]
    \centering
    \caption{Average accuracy for generating the entire output sequence correctly and its standard deviation in parentheses across five folds.}
    \begin{tabular}{|l|c|c|c|c|}
         \hline
         Validity Check & LSTM & Mean & BRNN & Transformer \\
         \hline
         Not checked & 0.125 (0.00672) & 0.118 (0.00459) & 0.130 (0.00677) & 0.113 (0.00495) \\
         \hline
         Checked & 0.139 (0.00552) & 0.122 (0.00389) & 0.140 (0.00503) & 0.123 (0.00324) \\
         \hline
    \end{tabular}
    \label{tab:entire-sequence-correct-results-table}
\end{table*}

\begin{table*}[!]
    \centering
    \caption{Average accuracy for individual code matching and its standard deviation in parentheses across five folds.}
    \begin{tabular}{|l|c|c|c|c|}
         \hline
         Validity Check & LSTM & Mean & BRNN & Transformer \\
         \hline
         Not checked & 0.749 (0.00111) & 0.742 (0.00167) & 0.751 (0.00219) & 0.730 (0.00303) \\
         \hline
         Checked & 0.764 (0.00339) & 0.742 (0.00145) & 0.765 (0.00264) & 0.742 (0.00210) \\
         \hline
    \end{tabular}
    \label{tab:single-code-correct-results-table}
\end{table*}

\begin{table*}[!]
    \centering
    \caption{Average accuracy for generating the underlying cause of death correctly and its standard deviation in parentheses across five folds.}
    \begin{tabular}{|l|c|c|c|c|}
         \hline
         Validity Check & LSTM & Mean & BRNN & Transformer \\
         \hline
         Not checked & 0.516 (0.00809) & 0.495 (0.00441) & 0.521 (0.0105) & 0.494 (0.00474) \\
         \hline
         Checked & 0.531 (0.00586) & 0.506 (0.00395) & 0.533 (0.00415) & 0.510 (0.00531) \\
         \hline
    \end{tabular}
    \label{tab:underlying-correct-results-table}
\end{table*}

Furthermore, we also include the results for the other three evaluation criteria. As shown in \textbf{Table \ref{tab:entire-sequence-correct-results-table}}, BRNN model achieves the best result in correctly generating the entire sequence with or without validity check. Notice that when evaluating a single pair of generated sequence and the target sequence, the output of this criteria is binary: 0 or 1. If any individual code fails to match, or the order of any code is incorrect, we will return a "0" for that pair of sequences. Thus, 0.140 indicates that BRNN model predict 14\% of sequences correctly.

As shown in \textbf{Table \ref{tab:single-code-correct-results-table}}, all four models achieve between 73\% and 76.5\% accuracy in predicting individual codes correctly. Comparing with the results in \textbf{Table \ref{tab:entire-sequence-correct-results-table}}, we argue that the encoder-decoders frameworks are much more accurate in finding the right codes on cause of death, than figuring out the right order between/among these codes.

Lastly, we show the accuracy in generating the underlying cause of death in \textbf{Table \ref{tab:underlying-correct-results-table}}. BRNN and LSTM models have better results than the other two, achieving over 53\% accuracy with validity check. Even though this is worse than the previous work using convolutional neural network (75\% accuracy), but the main objective of the previous work was to predict the underlying cause of death, which is obvious different from ours. Still, we would definitely try to improve the accuracy on underlying cause of death in future work.

CNN-based encoder-decoder framework frequently reported errors during the experiments. Conceptually, RNN-based encoder-decoder frameworks with attention mechanism are more suitable on sequence-to-sequence tasks; CNN-based model, on the other hand, finds it challenging to deal with sequences of different lengths. Consequently, we do not include the inconsistent results from CNN-based encoder-decoder framework.

\subsection{Visualization}

\begin{figure} [h!]
    \centering
    \includegraphics[scale=0.45]{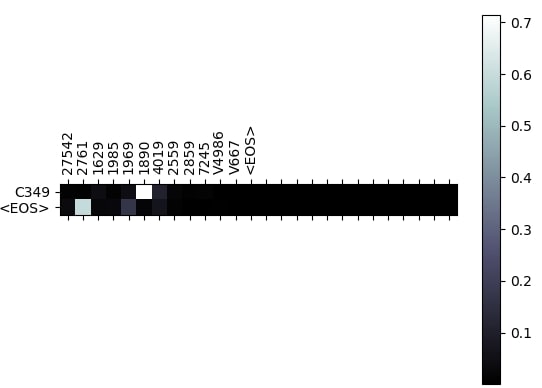}
    \caption{Visualization of attention in translation. The source sentence is on the top of the graph and the prediction is on the left. The special symbol $<EOS>$ is the end of sequence. The ICD-10 code $C349$ in human readable is malignant neoplasm of unspecified part of bronchus or lung; the ICD-9 code $1890$ is malignant neoplasm of kidney, except pelvis; the ICD-9 code $2761$ is hyposmolality and/or hyponatremia.}
    \label{fig:attention}
\end{figure}

In addition, we would like to show the visualization of attention during translation. As shown in \textbf{Figure. \ref{fig:attention}}, the source sentence is on the top of the graph while the predicted sentence is on the left. In this example, the code $C349$ (malignant neoplasm of unspecified part of bronchus or lung) is the correct prediction. We can observe that the code $1890$ (malignant neoplasm of kidney, except pelvis) from the source code has high attention with the predicted code. Yet the non-negligible flaw is that the source code $2761$ (hyposmolality and/or hyponatremia) has wrong attention with the $<EOS>$ symbol in the prediction.

\subsection{XLM}

To our surprise, the state-of-the-art algorithm XLM (cross-lingual language model) performs much worse than the other encoder-decoder frameworks. All BLEU scores are less than 1 after trying different combinations of hyper-parameters. Even though the pre-training can successfully finish after 500-700 epochs on monolingual corpora using masked language model perplexities as evaluation metric, the training on parallel data using BLEU score as evaluation metric fails to converge properly. 

We argue that the core algorithm behind BERT and XLM, masked language model, does not work on our data set. The idea of masked language modeling is to randomly mask (hide, making it unknown) a few words in the sentence (either source or target sentence) during the training stage and then to recover these masked words based on surrounding context. Since in average our target sentence has $2.25$ words, masking one word can make it extremely difficult to recover. Even worse, over $31\%$ of our target sentences consist of only one word: masking the only word makes it impossible to recover. 
\section{FHIR Interface}

\begin{figure}[h!]
    \centering
    \includegraphics[scale=0.45]{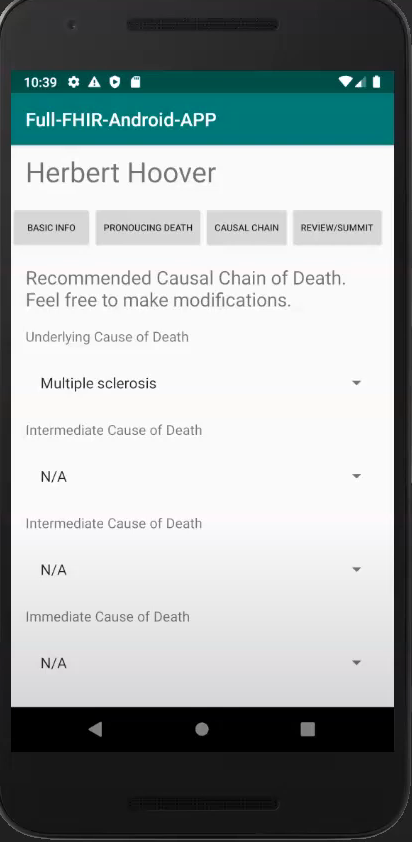}
    \caption{Screenshot of causal chain prediction display in the Android app. From top to bottom, the diagnosis codes correspond to the order of the output sentence. The ICD-10 codes are mapped into human-readable short descriptions.}
    \label{fig:causal-chain-display}
\end{figure}

We have implemented a prototype Android mobile application to demonstrate the usefulness and applicability of this work. This prototype app supports causal chain prediction, patient search, patient information display, determining causes of death, and review/submit. The app allows a physician to add death-related information when filling out the "Pronouncing Death" screen and the data bundle is compatible with FHIR servers. When generating causal chain of death for clinical decision support, the app will automatically retrieve medical condition codes from the FHIR server, and query the python pre-trained neural machine translation model for generating the casual chain of death. \textbf{Figure. \ref{fig:causal-chain-display}} shows a screenshot of the FHIR Android app when displaying the causal chain of death. The ICD-10 codes have already been mapped into human-readable short descriptions. From top to bottom, we show the ordered causes of death (from underlying cause of death to immediate causes of death). "N/A" indicates null value for the cause of death.

The use of mobile apps such as this prototype for delivering public health informatics creates the opportunity for real-time, point-of-care feedback. In the immediate term, clinicians completing mortality reporting data can be provided with decision support capability to improve the accuracy and completeness of the causes of death reported. In the future, such infrastructure may even be able to provide predicted causes of death for still-living patients, enabling predictive medical care.

\section{Conclusion}

In this paper, we are the first to successfully propose the causal chain of death using neural machine translation frameworks to support the timely, accurate and complete death reporting. The generated sequence has a BLEU score of $16.04$, close to the performance of the state-of-the-art in natural language domain (English-Czech translation task achieving BLEU score $17.7$ given the same vocabulary size around 10k). We also evaluate the model performance using three different accuracy scores, achieving 76.5\% accuracy in generating the individual code in output sequence.

In addition, we incorporate and explore medical domain knowledge as constraint when generating output sequence. We show that domain knowledge constraint has mixed impact on the encoder-decoder models, and argue that these models can learn the causal relationship between diagnosis codes from the data. Meanwhile, we demonstrate that validity check can be a critical step in the pipeline, which improves the stability of the encoder-decoder frameworks and may slightly improve the results.

Furthermore, we have visualized the results with attention mechanism, providing a tool to explore the relationship between certain condition codes in source sentence and those in target sentence. Lastly, we demonstrate a FHIR compatible mobile app to retrieve, modify and upload data.

Still, there are a few limitations with this work. The visualization of attention mechanism clearly shows the failure of alignment during translation, largely due to the extremely imbalanced lengths between source and target sentences. Furthermore, even though that the cross-lingual language modeling (XLM) has proven its effectiveness in natural language translation, it fails on our task. One potential cause is that the masked language modeling might not work on extremely short sentences (in average $2.25$ words per sentence).

One unsolvable problem is the one-word target sentence. Only in very rare condition we shall see a sentence consists of just one word in natural language; yet $31.77\%$ of training data, $31.68\%$ of validation data and $31.27\%$ of testing data are one-word target sentences. These samples significantly undermine the effectiveness of neural machine translation models.

Future work includes data augmentation so that the target sentence length will fit the newest masked language model (such as XLM). The model framework needs adapting towards the imbalanced source and target sentences.






\section*{Acknowledgment}
The author would like to thank Paula Braun (CDC) for her invaluable assistance and support in shaping this project. This article does not reflect the official policy or opinions of the CDC, NIH or NSF and does not constitute an endorsement of the individuals or their programs.


%





\ifCLASSOPTIONcaptionsoff
  \newpage
\fi





\bibliographystyle{IEEEtran}
\bibliography{IEEEabrv,Bibliography}

\vfill


\end{document}